\documentclass[10pt,twocolumn,letterpaper]{article}

\usepackage{cvpr}
\usepackage{times}
\usepackage{epsfig}
\usepackage{graphicx}
\usepackage{amsmath}
\usepackage{amssymb}
\usepackage{bm}
\usepackage{algorithm}
\usepackage{enumerate}
\usepackage{multirow}
\usepackage{array}
\usepackage{latexsym}
\usepackage{algpseudocode}
\usepackage{color}
\usepackage[pagebackref=true,breaklinks=true,letterpaper=true,colorlinks,bookmarks=false]{hyperref}

\cvprfinalcopy % *** Uncomment this line for the final submission

\def\cvprPaperID{1020} % *** Enter the CVPR Paper ID here

\ifcvprfinal\fi
\begin{document}

%%%%%%%%% TITLE
\title{Deep Transfer Learning for Person Re-identification}

\author{Mengyue Geng
\and
Yaowei Wang
\and
Tao Xiang
\and
Yonghong Tian
}

\maketitle
%\thispagestyle{empty}

%%%%%%%%% ABSTRACT
\begin{abstract}
\vspace{-0.3cm}
Person re-identification (Re-ID) poses a unique challenge to deep learning: how to learn a deep model with millions of parameters on a small training set of few or no labels.
In this paper, a number of deep transfer learning models are proposed to address the data sparsity problem. First, a deep network architecture is designed which differs from existing deep Re-ID models in that (a) it is more suitable for transferring representations learned from large image classification datasets, and (b) classification loss and verification loss are combined, each of which adopts a  different dropout strategy. Second, a two-stepped fine-tuning strategy is developed to transfer knowledge from auxiliary datasets. Third, given an unlabelled Re-ID dataset, a novel unsupervised deep transfer learning model is developed based on co-training. The proposed models outperform the state-of-the-art deep Re-ID models by large margins: we achieve Rank-1 accuracy of 85.4\%, 83.7\% and 56.3\% on CUHK03, Market1501, and VIPeR respectively, whilst on VIPeR, our unsupervised model (45.1\%) beats most supervised models.

\end{abstract}

\vspace{-0.6cm}
%%%%%%%%% BODY TEXT
\section{Introduction}
\label{sec:introduction}
\vspace{-0.2cm}
Person re-identification (Re-ID) is the problem of matching people across non-overlapping
camera views, which  typically arises in a surveillance application. Despite the best efforts from the computer vision
researchers, it remains an unsolved problem \cite{zheng2016person}.
Earlier works focus on either designing view-insensitive feature representations \cite{farenzena2010person,yang2014salient,kviatkovsky2013color,ma2012local,zhao2014learning,matsukawa2016hierarchical}, or learning an effective distance metric \cite{xiong2014person,ma2014person,lisanti2015person,chen2016similarity,zhang2016learning,zhangsample,martinel2016temporal}, or both  \cite{gray2008viewpoint,liao2015person}. Recently, inspired by the success of deep neural networks, particularly deep Convoluational Neural Networks (CNNs) in various vision problems \cite{krizhevsky2012imagenet,simonyan2014very,szegedy2015going,he2015deep}, deep Re-ID models started to attract attention \cite{li2014deepreid,yi2014deep,
ding2015deep,ahmed2015improved,
ShiZLLYL15,UstinovaGL15,varior2016gated,liu2016end,cheng2016person,xiao2016learning}.

However, unlike other visual recognition problems, especially closely related ones such as face verification, only limited success has been achieved so far by deep Re-ID models: they only marginally improve over the hand-crafted feature + metric learning based alternatives on large datasets such as Market1501 \cite{zheng2015scalable}, and are outperformed on small datasets such as VIPeR \cite{gray2007evaluating}. Lack of large labelled training set is an obvious reason. Collecting matching pairs of person images in a camera network is a notoriously difficult task \cite{zhao2013unsupervised}. As a result, even the largest published Re-ID datasets only have modest sizes: 1,360 unique identities in CUHK03 \cite{li2014deepreid} and 1,501 in Market1501 \cite{zheng2015scalable}. In contrast, the widely used LFW dataset \cite{LFWTech} for face verification has 5,749 identities -- faces of celebrities are much easier to collect and label than passers-by captured by a surveillance camera network. Importantly, one could easily collect a much larger auxiliary dataset of faces to assist in the model learning: one of the state-of-the-art results on LFW was obtained by pretraining the deep model on an auxiliary face dataset of 200M images of 8M identities \cite{FaceNet_cvpr15}.

Given insufficient training samples, transferring feature representations learned from a larger auxiliary dataset becomes necessary. Indeed, transfer learning has been considered in most existing deep Re-ID works. In particular, given a small Re-ID dataset with only a few hundreds of labelled identities, existing models typically pretrain with larger Re-ID datasets followed by fine-tuning on the target set, with a notable exception of \cite{xiao2016learning} which learns  a single model jointly across multiple Re-ID datasets before the fine-tuning in each. In other words, only Re-ID datasets are considered as auxiliary datasets -- hardly ideal because all Re-ID datasets published so far are relatively small. Importantly the domain gaps between different Re-ID datasets are often large due to the drastically different camera viewing conditions; designing the most suitable method to prevent negative transfer is thus a challenging task  \cite{pengunsupervised}.

We argue that to transfer knowledge that is generalisable to any Re-ID dataset, we should go beyond existing Re-ID datasets and consider much larger sources. An obvious choice would be the ImageNet dataset \cite{deng2009imagenet} which contains millions of images of thousands of object categories and has been shown to be useful as an auxiliary dataset for model pretraining for a variety of visual recognition tasks \cite{DeCAF}. However, transferring  knowledge from ImageNet to a Re-ID dataset has a number obstacles. First, the object categorisation task of ImageNet is very different from the object instance verification task of person Re-ID. Second, the inputs to a Re-ID model are person detection images in CCTV surveillance videos, which have very different aspect ratios (people are typically upright) and much lower resolutions. This is the reason why most recent deep Re-ID models \cite{xiao2016learning,
cheng2016person,varior2016gated,liu2016end,ShiYZLLZL_ECCV16} have very different network architectures compared with those of the models excelled on the ImageNet object categorisation task, e.g.~having smaller filter size, and being shallower with most pooling layers removed. Such models are designed for training from scratch on Re-ID datasets, and are unsuitable for knowledge transfer from ImageNet.

The proposed deep Re-ID network architecture in this work is designed specifically for transferring generalisable feature representations learned from ImageNet to Re-ID datasets. To this end, we make two key design choices: (1) The base network structure is a standard GoogleNet \cite{szegedy2015going} which has been optimised for ImageNet. (2) Two losses are combined. These include an identity classification loss which is chosen because the model needs to be pretrained on the auxiliary ImageNet dataset for the object classification task. The other loss is a verification loss, which aims to learn a feature representation for matching person. We argue that by combining the two losses, our model can bridge the large task discrepancy between object categorisation and object instance verification, as well as the large domain gap between ImageNet and Re-ID datasets.

Apart from having different architecture and training objectives compared to existing deep Re-ID models, the proposed model also has the following distinctive features. (1) Since all target Re-ID datasets are relatively small given millions of model parameters, avoiding overfitting is of paramount importance. Dropout is a widely adopted technique for overcoming overfitting. In our model, two different dropout strategies are employed for the two different losses. (2) We propose a two-stepped fine-tuning strategy after the model is pretrained due to the unique combination of the two losses. Compared to the conventional one-stepped fine-tuning strategy, it is much more effective as shown in our experiments (see Sec.~\ref{sec:experiment:supervised}).

%In general both source and target are unlabeled in unsupervised transfer learning.
Learning a Re-ID model given a set of unlabelled data is a more challenging task, but also has more practical uses in real-world applications -- a Re-ID  system is typically installed for a camera network monitoring a large public space (e.g.~a train station or a shopping mall) which can easily consist of hundreds of cameras. Even labelling a few hundreds people across all camera views is infeasible. However, person detection images can be readily obtained in each view using a person detector, resulting in an unlabelled Re-ID dataset. Transfer learning from labelled source data to unlabelled target data is an unsupervised\footnote{By `unsupervised', we mean target-unsupervised domain adaptation, a definition adopted by \cite{Ganin_ICML15,DBLP:journals/corr/ZhangYCW15,DBLP:journals/corr/Long0J16}.}domain adaptation problem which has not been studied by existing deep Re-ID models. In this work, a novel co-training based unsupervised transfer learning model is proposed. Specifically, the model alternates between a graph regularised discriminative dictionary learning model and a soft-label self-training deep model, with the former providing the soft-labels  for the latter and the graph regularisation provided in the opposite direction. We show that such a deep/non-deep hybrid co-training framework can effectively prevent model drift and yield Re-ID performance that is better than most existing supervised learning based models.

%------------------------------------------------------------------------
\vspace{-0.3cm}
\section{Related Work}
\label{sec:relatedworks}
\vspace{-0.2cm}

\noindent \textbf{Deep Re-ID model}\quad
Existing deep Re-ID models \cite{li2014deepreid,yi2014deep,ding2015deep,
ahmed2015improved,ShiZLLYL15,UstinovaGL15,
varior2016gated,
liu2016end,cheng2016person,xiao2016learning,ShiYZLLZL_ECCV16} differ significantly in their network architectures, which are largely determined by the training objectives/losses. Specifically, most existing works cast the Re-ID problem as a deep metric learning problem and employ  pairwise verification loss \cite{yi2014deep,ahmed2015improved,ShiZLLYL15,UstinovaGL15,
varior2016gated,ShiYZLLZL_ECCV16} or triplet ranking loss \cite{ding2015deep,liu2016end,cheng2016person}, or both \cite{wangjoint}. Correspondingly the overall network architecture is a Siamese CNN network with either two or three branches for the pairwise or triplet loss respectively. None of them uses an identity classification loss with the only exception of \cite{xiao2016learning} which has an one-branch architecture. In contrast, our model has a Siamese two-branch architecture with an identity classification loss for each branch and pairwise verification loss across the two branches. This architecture is similar to the one used for deep face verification \cite{Deep_Face_joint_NIPS2014}. Combining the two loses in   \cite{Deep_Face_joint_NIPS2014} aims to exploit the strengths of the two losses: the classification loss pulls different classes apart and the verification loss makes the intra-class distance small. In contrast, we choose the combination so that the model can be pretrained on the ImageNet object classification task -- among the two losses the classification loss makes sure that the ImageNet-learned representation is relevant whilst the verification loss guides the adaptation towards the person Re-ID dataset/verification task.

Apart from the overall architecture (one, two, or three branches), existing models also have very different base network structure (the convolution/pooling layers in each branch). It is noted that most recently proposed  deep Re-ID models \cite{wangjoint,varior2016gated,liu2016end,cheng2016person,
xiao2016learning,ShiYZLLZL_ECCV16} have base networks tailor-made for the Re-ID problem, that is, they  take into consideration the smaller input image size and  different (non-square) aspect ratio of person detection images in a Re-ID dataset. In particular, the filter-size/stride step are typically much smaller with fewer pooling layers, compared to the ImageNet-oriented GoogleNet \cite{szegedy2015going} or VGG Net \cite{simonyan2014very}, so that they can be learned from scratch using Re-ID datasets alone. However, this simplified base network architecture, together with the lack of the classification-verification loss combination mean that the existing models are unable to exploit the rich transferable feature representation learned from ImageNet.

\noindent \textbf{Dropout strategy}\quad
Dropout \cite{srivastava2014dropout} is a widely adopted technique in deep learning to counter overfitting, a problem that is particularly acute in Re-ID due to the small data size. Given the two losses, we propose to use different dropout strategies for each loss-associated layers. Specifically, the standard random dropout \cite{srivastava2014dropout} is deployed for the classification loss layers, whilst for the pairwise verification loss layers, we introduce pairwise-consistent dropout, that is, each pair of compared training data points share the same dropout mask.
% This is intuitive: the same set of units should be dropped if two feature maps are to be compared.
We show experimentally that such a modification can bring about 3\% improvement in Re-ID accuracy. %After model training, existing works also differ in how the learned model is used for testing, i.e.~matching a probe image against a set of gallery images. This is determined by whether the model aims to learn a single-image representation (SIR) or cross-image representation (CIR) \cite{wangjoint}. If CIR is used for matching, the testing speed would be a few magnitude slower than the SIR learning only models. This is because with CIR, each gallery image needs to be fused together with the probe to compute CIR whilst the SIR can be pre-computed and matching is done by simple Euclidean distance calculation. Despite using a two-branch architecture with pairwise loss, our model aims to learn SIR only, and is thus much more efficient during testing and hence more suitable for real-time applications than the CIR based models \cite{li2014deepreid,wangjoint,ShiZLLYL15,varior2016gated,liu2016end,ShiYZLLZL_ECCV16} whilst yielding much higher matching accuracy.

\noindent \textbf{Deep transfer learning}\quad
Transfer learning or domain adaptation is an extensively studied topic \cite{transfer_survey}. Transfer learning is widely used for deep learning when a target task is short of labelled data.  The most common deep transfer learning strategy is fine-tuning \cite{How_transferrable_NIPS2014}: first train a base network using a large source data and then copy the first $n$ layers to the corresponding layers of the target network, followed by randomly initialising the remaining layers and finally fine-tune only them or all layers. A systematic study is presented in \cite{How_transferrable_NIPS2014} which examines how transferable features of different layers are between the source and target domains. It concludes that the generalisation ability diminishes when the discrepancy between the source and target tasks increases. Note that the source and target tasks considered in \cite{How_transferrable_NIPS2014} were classifying different subsets of ImageNet, so the task/domain discrepancy studied is nowhere near as big as in our ImageNet $\rightarrow$ Re-ID transfer setting. As a result, the conventional one-stepped fine-tuning strategy becomes inadequate. To overcome the large task discrepancy between classification and verification, we propose a two-stepped fine-tuning strategy whereby  the network is first fine-tuned with the classification loss only, followed by fine-tuning with both classification and verification  losses.

Note that beyond fine-tuning, several recent works take a multi-task joint training approach \cite{Long_ICML15,Ganin_ICML15,DBLP:journals/corr/ZhangYCW15,DBLP:journals/corr/Long0J16a,DBLP:journals/corr/Long0J16,xiao2016learning}, one of which is designed specifically for Re-ID \cite{xiao2016learning}. Most of them aim to minimise the discrepancy between the marginal \cite{Long_ICML15,Ganin_ICML15,DBLP:journals/corr/ZhangYCW15} or joint \cite{DBLP:journals/corr/Long0J16a} distributions of the source and target data, e.g., by introducing a cross-domain loss that is designed to blur the domain boundary \cite{Ganin_ICML15}. However, these works  assume that the tasks are the same or similar in the two domains, e.g., classifying the same object classes shared by two datasets. They are thus  unsuitable when the source and target domains have completely different tasks, in our case object categorisation in ImageNet and person matching in Re-ID -- aligning the data distributions of the two datasets would not make any sense. The joint learning + multi-task learning + fine-tuning based deep Re-ID model in \cite{xiao2016learning} is clearly not suitable for transferring from ImageNet to Re-ID with the different source and target tasks.

\begin{figure*}[ht!]
	\centering
	\setlength{\belowcaptionskip}{-50pt}
	\setlength{\abovecaptionskip}{-50pt}
	\includegraphics[width = 0.7\linewidth]{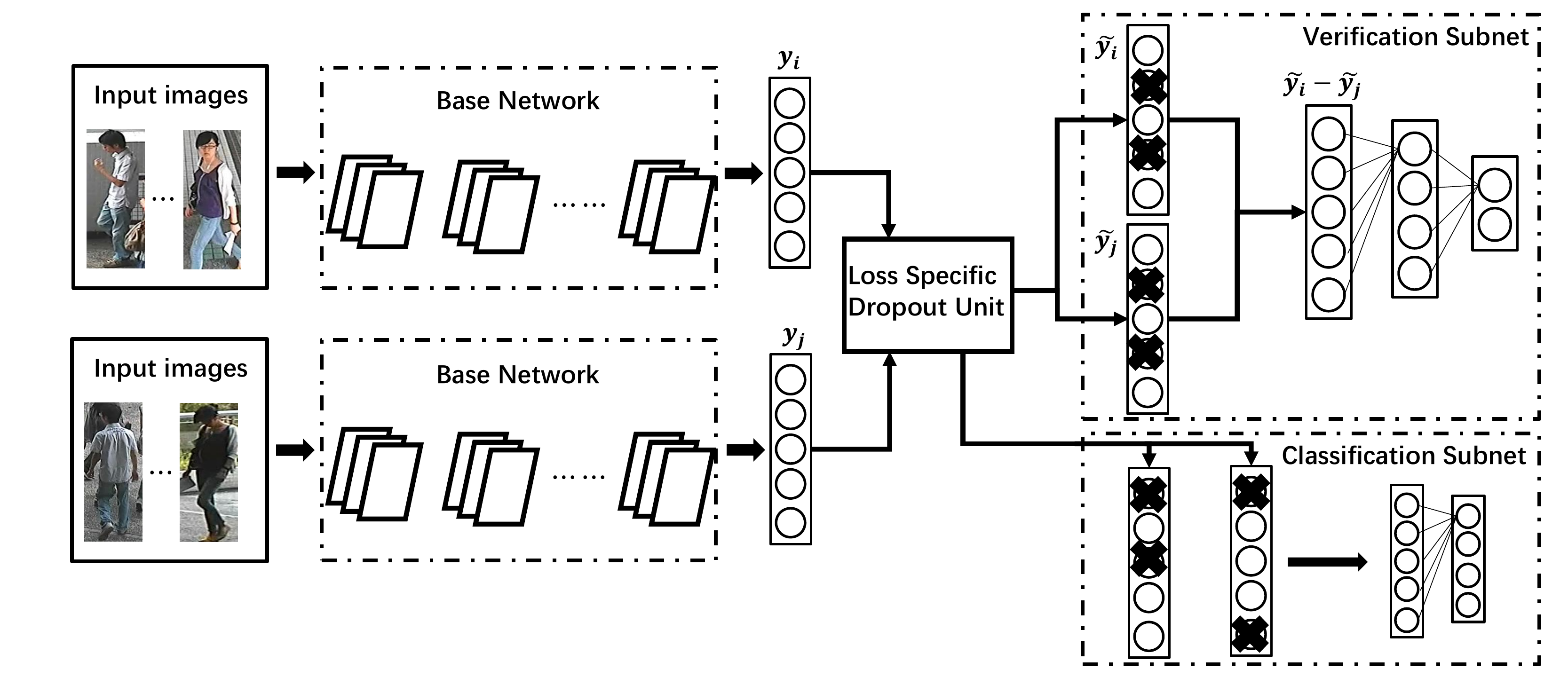}
	\caption{The proposed deep Re-ID network architecture. }	\label{fig:the_network}
\end{figure*}
\noindent \textbf{Deep unsupervised domain adaptation}\quad
%co-training is well studied. however, now between a deep and no-deep model.
 In theory, any unsupervised deep learning methods can potentially be applied for domain adaptation when the first $n$ layers are pretrained on the source data. These include auto-encoder \cite{hinton2006reducing} and dictionary learning \cite{Kreutz-DelgadosDictionary} which can be implemented as neural network layers and integrated as the later/top layers of a CNN network \cite{Wang_2016_CVPR}. The main limitation of an unsupervised model is that it cannot learn discriminative features. Hence soft-label self-training based deep unsupervised learning has become popular recently \cite{Huang_CVPR16}. In this work, a novel co-training \cite{Nigam:2000:co-training} based unsupervised domain adaptation method is proposed to overcome the main drawback of self-training based methods, i.e., model drift \cite{siva2011weakly}. Combined with the proposed two-stepped fine-tuning method, this gives us a powerful deep unsupervised Re-ID model that outperforms not only alternative unsupervised models, but also most supervised models which use training labels. Recently a number of deep unsupervised transfer learning models are proposed  \cite{Ganin_ICML15,DBLP:journals/corr/ZhangYCW15,DBLP:journals/corr/Long0J16}. Nevertheless, the domain gap between different Re-ID datasets is significant and cannot be overcome by just aligning the data distributions, making them less effective than the proposed co-training based unsupervised learning method, as demonstrated in our experiments.

\noindent \textbf{Our contributions} are as follows:
(1) A new deep Re-ID network architecture is designed for transferring feature representation learned  from large image classification datasets. It is unique in its loss combination and dropout strategy. (2) A two-stepped fine-tuning strategy is further developed for deep transfer learning. (3) A novel co-training based unsupervised domain adaptation method is proposed for unsupervised Re-ID. (4) We present comprehensive experimental evaluations on 5 benchmarks. Our experiments show that the proposed models outperform the state-of-the-art deep Re-ID models by a significant margin. %we achieved Rank-1 accuracy of 85.4\%, 83.7\% and 56.3\% on CUHK03, Market1501, and VIPeR respectively, whilst on VIPeR, our unsupervised model (45.1\%) beats most supervised alternatives.

\vspace{-0.3cm}
%------------------------------------------------------------------------
\section{Deep Re-ID Model}
\label{sec:model}
\vspace{-0.2cm}
\subsection{Network Architecture}
\label{sec:architecture}
\vspace{-0.2cm}

\noindent \textbf{Overview} The overall network architecture of the proposed deep Re-ID model is illustrated in Fig.~\ref{fig:the_network}. It is essentially a two-branch Siamese  network that takes a pair of input person detection images as input and aims to learn a deep representation of person appearance that is identity-discriminiative  so that images of the same person can be matched correctly whilst visually similar people can be distinguished. The model has two training tasks/objectives/losses: an ID classification loss and a pairwise verification loss. As a result, the network contains four parts (see Fig.~\ref{fig:the_network}): a base network shared by the two branches, a loss-specific dropout unit, an ID classification subnet, and  a pairwise verification subnet. The two main branches of the network have the same base network architecture and share their parameters, hence the name Siamese.  After feature vectors are computed for the input images using the base network, they are fed into a loss-specific dropout unit so that either a pairwise-consistent dropout or the standard random dropout is applied to the features. After that, the pairwise verification subnet takes a pair of features and learn to distinguish whether they come from the same person or not. In the meantime, the person ID classification subnet learns to classify each feature output of the base network into a class corresponding to the input image person ID.

\noindent \textbf{Base network}\quad The base network is a CNN that learns a deep representation from the input images. Various CNN architectures can be considered. In this paper we use GoogLeNet \cite{szegedy2015going}. This is different from the base networks used by  most existing deep Re-ID models. We deliberately choose an existing network that is competitive in the ImageNet classification benchmark, and has been widely used in many other vision problems, rather than designing a simplified bespoke Re-ID network. This is because we aim to use the network to transfer generalisable feature representations from the much larger ImageNet dataset. Among the recently proposed networks that achieved good classification performance on ImageNet, GoogLeNet is chosen over  VGG net \cite{simonyan2014very}  due to the fact that it has much smaller parameter size\footnote{We found that the residual networks \cite{he2015deep} have a similar performance as GoogLeNet when used as the base network in our model.}. %We show in our experiments that this base network is much more effective than an alternative Re-ID specific base network in our transfer learning model.

\noindent \textbf{Loss specific dropout unit}\quad Given an input image $\textbf{x}$,  the base network produces a $D$-dimensional vector of output
$\textbf{y}$. It will then enter a loss specific dropout unit where the operation depends on whether the output of the unit $\widetilde{\textbf{y}}$ is fed into the classification subnet subject to the classification loss or the verification subnet with a pairwise verification loss. For the former, the standard dropout operation \cite{srivastava2014dropout} takes place. Concretely it will
randomly set part of the elements of $\textbf{y}$ to zero. Formally, we have
\vspace{-0.3cm}
\begin{equation}
\begin{aligned}
\widetilde{\textbf{y}} = \textbf{r}* \textbf{y}
\end{aligned}
\end{equation}
where the $D$-dimensional vector $\textbf{r}$ is a dropout mask and $*$ denotes an element-wise product.  Each element of $\textbf{r}$ is a random variable sampled following a Bernoulli process, i.e.,~ the $d$-th element $r_d \sim Bernoulli(p)$ and has a probability of $p$ to be $1$. Given a different image in a mini-batch, a different random dropout mask will be applied and the output is then fed into the classification subnet.

Such a completely random dropout operation is nevertheless not appropriate when the output $\widetilde{\textbf{y}}$ is to be fed into the verification subnet. In particular, as to be detailed later, given a mini-batch of training images, each pair of images will have their  output vectors $\widetilde{\textbf{y}}_i$ and $\widetilde{\textbf{y}}_j$ subject to an element-wise subtraction operation in the subnet. Therefore, if two randomly generated dropout masks $\textbf{r}_i$ and $\textbf{r}_j$ are applied, the difference of two vectors could be caused by the different random masks rather than the appearances of the two compared people -- a clearly undesirable effect. To address this problem, we introduce a pairwise-consistent dropout for the verification subnet, that is, we make sure that $\textbf{r}_i=\textbf{r}_j$ when the $i$-th and $j$-th images are compared in the verification subnet.  Interestingly, although it is intuitive, we could not find any published work that discuss this need to design different dropout strategy for layers used as input to pairwise or triplet losses. For example, all relevant codes we could find on GitHub use the standard random dropout. We show in our experiments that by adopting the pairwise-consistent dropout for the verification loss, a 3\% improvement can be obtained compared to the random dropout.

\noindent \textbf{Person ID classification subnet} The person ID classification part learns a softmax classifier with a cross-entropy loss that distinguishes different people from each other. After the features are extracted from the base network and the random dropout is applied, a softmax layer with $N$ nodes are then connected, where $N$ is the unique person number in the training set.

\noindent \textbf{Pairwise verification subnet} The pairwise verification subnet first takes two feature vectors $\widetilde{\textbf{y}}_i$ and $\widetilde{\textbf{y}}_j$ as input. They are first fused with element-wise subtraction. Subsequently, the difference vector is passed to a rectified linear unit (ReLU).
After a fully connected (FC) layer, the last layer of the verification network is a softmax layer with two output nodes, corresponding to whether or not the input image pair contains the same person. Note that for pairwise verification, the margin based contrastive loss is much widely used beyond Re-ID \cite{Deep_Face_joint_NIPS2014}. However, for Re-ID, with a few exceptions \cite{varior2016gated,varior2016siamese}, this subtraction + binary cross-entropy loss is more popular \cite{wangjoint,yi2014deep,UstinovaGL15}. We find empirically that using the contrastive loss leads to worse performance in our model. It is also worth pointing out that more sophisticated differencing operations have been developed to deal with the mis-alignment issue of compared person detection images \cite{li2014deepreid,ahmed2015improved}; however they cannot be applied this late in our network architecture after a forward-pass of the GoogLeNet base network. A full-blown Mahalanobis metric learning loss \cite{ShiYZLLZL_ECCV16} could also be deployed in our verification subnet, but that will sacrifice the testing efficiency as the model cannot be used as SIR (single image representation) \cite{wangjoint} model  any more.    %The detailed network structure for the two subnets, as well as how the training data is organised in a mini-batch to form pairs for the verification subnet can be found in the supplementary material.

%\begin{figure*}[ht!]
%	\centering
%\setlength{\belowcaptionskip}{-50pt}
%	\setlength{\abovecaptionskip}{-50pt}
%	\includegraphics[width = 0.8\linewidth]{two_staged_finetuning.pdf}
%	\caption{Illustration of the proposed two-stepped fine-tuning strategy.}
%	\label{fig:two_staged_finetuning}
%\end{figure*}
\vspace{-0.2cm}
\subsection{Model Training and Testing}
\label{sec:testing}
\vspace{-0.2cm}
Our network is too big to train effectively from scratch using existing person Re-ID datasets. Transfer learning using other datasets as auxiliary data is thus necessary. The deep transfer learning models developed in this paper will be described in the next section. Here we focus on the testing part, that is, after the model is learned, how to use it for matching a probe image against a set of gallery images in a test set. Note that since the test people have different identities as the training people, the ID classification subnet is redundant during testing. The verification subnet could potentially be used to generate a matching score -- given the probe image and each gallery image, they can be fed into the network to compute the same-identity/different-identity score. However, by doing so the model becomes a cross image representation (CIR)  model \cite{wangjoint}, which means that the input image pair have to go through the FC layer in the subnet and the softmax loss layer. Instead, we intend to use our model as a SIR model, that is, we pre-compute the output vector of the base network $\textbf{y}$ for the gallery; and when any probe comes in, we compute its feature output and compare with the gallery output vectors using a simple Euclidean distance, which is about 3 magnitude faster in our model than entering the verification subnet and computing the softmax score as the distance. This testing procedure is clearly more suitable for real-time applications than those of the alternative CIR models \cite{li2014deepreid,wangjoint,ShiZLLYL15,varior2016gated,
liu2016end,ShiYZLLZL_ECCV16}.

\vspace{-0.3cm}
\section{Deep Transfer Learning for Re-ID}
\label{sec:crossdomain}
\vspace{-0.2cm}

% first mention the three transfer learning scenarios and focus the large to smaller which also have two scenario: with and withhout label.
We consider two transfer learning settings: supervised when the target Re-ID dataset is labelled with person identities, and unsupervised when it is unlabelled.

\vspace{-0.2cm}
\subsection{Supervised Transfer Learning}
\label{sec:crossdomain:supervised}
\vspace{-0.1cm}

\noindent \textbf{Staged transfer learning} \quad As in existing Re-ID works, there are two scenarios under the supervised setting: the  target Re-ID dataset is `large', i.e.~having more than 1,000 identities, for instance CUHK03 \cite{li2014deepreid} and Market1501 \cite{zheng2015scalable}, and it is `small'  with less than 1,000, e.g.~VIPeR \cite{gray2007evaluating}. Existing deep Re-ID models are trained from scratch for the large datasets, i.e., without transfer learning. For the small datasets, the models are typically pretrained on large datasets (e.g., CUHK03+Market1501), and then fine-tuned on the small target dataset. We call this an one-staged transfer learning method based on an one-stepped fine-tuning strategy.

With the unique combination of classification and verification losses and the corresponding two subnets, transfer learning from ImageNet is conducted for our model regardless of the target dataset size. Specifically, for a large Re-ID dataset, the transfer learning is one-staged, i.e., ImageNet $\rightarrow$ Re-ID dataset, whilst two-staged transfer learning is required when the target dataset size is small, i.e., ImageNet $\rightarrow$ large Re-ID datasets $\rightarrow$ small Re-ID dataset. Importantly, in each stage, we develop a two-stepped fine-tuning strategy for more effective transfer learning compared with the conventional one-stepped one.

\noindent \textbf{Two-stepped fine-tuning} \quad This strategy is described based on the second stage of the small dataset scenario, i.e.~large Re-ID datasets $\rightarrow$ small Re-ID dataset. The same strategy is adopted for the ImageNet $\rightarrow$ large Re-ID dataset transfer learning.

%Training a deep network with millions of parameters on a small Re-ID dataset with hundreds of images is a tall order. Consequently, it is necessary to transfer knowledge from a large source dataset to the small target dataset. Going beyond the pre-training + fine-tuning based naive transfer widely adopted in previous works, we develop a staged model training strategy whereby different parts of the network are pre-trained with source and target data separately before the final stage of joint training takes place. Critically, the softmax ID classification layer is pretrained first to make sure that it is in harmony with the network main branch before it can help learn the target domain Re-ID model. We found that the staged training strategy can effectively avoid harmful gradients being backpropagated to the main branch.

Suppose we have a large source Re-ID dataset\footnote{If more than one are used, they are simply merged into one.} $S$ and a small target dataset $T$ with $N_s$ and $N_t$ unique person identities respectively. Given an initial model trained using  $S$, our goal is to transfer the learned feature representation from $S$ to $T$. Note that the softmax ID classification layer in the initial network cannot be re-used because the $N_s$ and $N_t$ identities have no overlap. The original $N_s$-nodes softmax layer thus has to be replaced with a randomly initialised one with $N_t$ nodes. In the first step of the fine-tuning, we freeze all other layers and train only the newly added softmax layer, i.e., the classification subnet. Freezing the other parts of network (base network + verification subnet) is critical for this stage of training: without locking them, the randomly initialised parameters of the softmax layer will backpropagate harmful gradients to the base network, generating `garbage gradients' that will derail the model adaptation.
After the softmax layer is fully trained so that the learned features from $S$ can do a decent job in classifying the $N_t$ new identities, in the second stage we fine-tune the softmax layer as well as all other layers of the network using the target dataset $T$.  We will show in our experiments (see Sec.~\ref{sec:experiment:supervised}) that the proposed two-stepped fine-tuning strategy is much better than an one-stepped one.

\vspace{-0.3cm}
\subsection{Unsupervised Transfer Learning}
\label{sec:crossdomain:unsupervised}
\vspace{-0.2cm}
Now the $M_t$ target training images of a unknown number of identities are unlabelled. For simplicity of symbols, we assume they are collected from two camera views denoted as $A$ and $B$ respectively.  Let's denote the training set as $\textbf{X} = \{\textbf{X}^a, \textbf{X}^b\}$, where $\textbf{X}^a = \{\textbf{x}_1^a,...,\textbf{x}_{M_a}^a\}$ contains  $M_a$ images in view $A$, while $\textbf{X}^b = \{\textbf{x}_1^b,...,\textbf{x}_{M_b}^b\}$ for the $M_b$ images in view $B$, we thus have $M_t = M_a + M_b$. For each image $\textbf{x}$, an $D$-dimensional feature vector $\textbf{y}=\phi(\textbf{x})$ is computed by the base network to represent its appearance, where $\phi$ denote the mapping function learned by the base network using the source dataset $S$. We wish to learn a better network using $T$ with $M_t$ unlabelled images yielding an updated mapping function $\widetilde{\phi}$.
%so that given $x_a$ and $x_b$ representing two test person images from $A$ and $B$ respectively, $||\widetilde{\phi}(x_a) - \widetilde{\phi}(x_b)||_F$ can be used for matching their identities.

\noindent \textbf{Self-training}\quad
One solution to the unsupervised transfer learning problem is to use the same two-staged supervised transfer learning model with two-stepped fine-tuning. Instead of using the identity labels to set the training objectives, we use soft (pseudo) labels. Specifically, for each of the $M_a$ images from camera $A$ $\textbf{x}_i^a \in \textbf{X}^a$, we assign it with a unique class label. After that, each of the $M_b$ images from camera $B$ is assigned with the same label as its nearest neighbour from $A$ based on $||{\phi}(\textbf{x}_i^a) - {\phi}(\textbf{x}_j^b)||_2$. Note that these labels clearly do not correspond to the real identity labels: for a start, there could be multiple images per person in each camera, so there are less than $M_a$ identities; second, the nearest neighbour can only give a visually similar person which by no means is always the same person. These soft labels are thus highly noisy. In a self-training strategy, the fine-tuned network will produce an updated mapping function $\widetilde{\phi}$ which will be used to generate another set of soft labels for retraining. Model drift is thus a big problem: the errors in the soft labels will be propagated with the iterations and quickly magnified.

\noindent \textbf{Co-training} \quad One solution to the model drift problem is co-training \cite{colt98/blum,Wang_co-training_ECML07}. It was first designed for using the same model with two sufficient and yet conditionally independent views (feature representations) as inputs to label some unlabelled instances for each other \cite{colt98/blum}. Since in most problem settings, such views do not exist, in practice one often has a co-training style algorithm whereby two different models with the same features or even same model with same feature but different parameter settings are used \cite{Wang_co-training_ECML07}. The key is that both models need to be somewhat effective and importantly complementary to each other.

In our case, we have already got the self-training deep CNN as one of the two models. The other unsupervised model needs to be of similar effectiveness yet complementary. To this end, we choose a graph regularised subspace learning model \cite{hu-et-al:hu2014smooth,yin-et-al:yin2015laplacian}. Such a model aims to learn a discriminative subspace where the data distribution is smooth with regard to a K-nearest neighbour (KNN) graph constructed in the input feature space. In such a learned subspace, data clusters can be formed to provide the soft-labels for the self-training deep model. In the meantime, it uses the deep model learned feature vector $\textbf{y}=\phi(\textbf{x})$ as model input as well as to construct the graph for regularisation.

Formally, given our pretrained deep Re-ID model, we obtain a feature matrix from the base network output $\mathbf Y = [\mathbf Y^{a},~ \mathbf Y^{b}] \in
\mathbb{R}^{D \times M_t}$,  where $\mathbf Y^{a} = [\mathbf y_1^{a},~...~, \mathbf y_{M_a}^{a}]
\in \mathbb{R}^{D \times M_a}$ and $\mathbf Y^{b} = [\mathbf y_1^{b},~...~,
  \mathbf y_{M_b}^{b}]\in \mathbb{R}^{D \times M_b}$. We aim to learn a subspace defined by a dictionary $\mathbf D$ and a new representation $\mathbf Z$ in the subspace.
$\mathbf D$ and $\mathbf Z$ can be estimated jointly by
solving the following optimisation problem:
\vspace{-0.3cm}
\begin{equation}
\label{eq:standardSC}
(\mathbf  D^{*}, \mathbf Z^{*}) = \min_{\mathbf D,\mathbf Z}    \| \mathbf Y- \mathbf{DZ}\|_F^2+\lambda \mathrm{\Omega(\mathbf Z)} ~~s.t.~~\|\mathbf{d}_i\|_2^2\le 1,
\end{equation}
where the first term is the reconstruction error
evaluating how well a linear combination of the learned atoms can
approximate the input data, and $||.||_F$ denotes the matrix
Frobenious norm. $\mathrm{\Omega(\mathbf Y)}$ is the graph regularisation term that is weighted by $\lambda$:
\vspace{-0.3cm}
\begin{equation}
\vspace{-0.1cm}
	\mathrm{\Omega(\mathbf{Z})}=\sum_{ij}^{}W_{ij}\|\mathbf{z}_i-\mathbf{z}_j\|_2^2.
	\label{eq: GL1}
\end{equation}
where the graph is encoded by an
affinity matrix $\mathbf{W}\in \Bbb{R}^{M_t\times M_t}$ for $M_t$ data
points where $W_{i,j}\neq 0$ only when $\mathbf{y}_i$ and $\mathbf{y}_j$ are from two different camera views and are nearest neighbours.  With the learned new representation  $\mathbf{Z}$, we can generate soft labels for the unlabelled target data, that is, the cross-view nearest neighbours are obtained by $||\textbf{z}_i^a - \textbf{z}_j^b||_2$ instead of  $||{\phi}(\textbf{x}_i^a) - {\phi}(\textbf{x}_j^b)||_2$. With these soft-labels, another round of self-training of the deep model is carried out and the updated base network then produces input vectors and new graph for the subspace learning model. This iterative process normally converges after 2-3 iterations.

\vspace{-0.3cm}
\section{Experiments}
\label{Sec:Exp}

\vspace{-0.2cm}
\subsection{Datasets and Settings}
\label{sec:experiment:dataset}
\vspace{-0.1cm}
\noindent \textbf{Datasets\quad} Five widely used datasets are used including two large datasets and three small ones.
\textbf{CUHK03} \cite{li2014deepreid}
contains 13,164 images of 1,360 identities from 6 cameras. We use the 20 standard training/test splits as provided in \cite{li2014deepreid}: 100 identities are randomly selected for testing and another 100 for validation, whilst the remaining 1160  for training. Both manually cropped and automated detected person images are used for our evaluations.
As in most previous works, we adopt the single-shot setting.
\textbf{Market1501} \cite{zheng2015scalable} contains 32,668 detected person bounding boxes of
1,501 identities from 6 cameras.  We use the
training and test splits provided in  \cite{zheng2015scalable} under both the single-query (SQ) and
multi-query (MQ) evaluation settings.
\textbf{VIPeR} \cite{gray2007evaluating}
contains 632 identities and each has two images
in two views with distinct view
angles.  The
632 identities are randomly divided into two equal
halves, one for training and the other for testing. The training process is repeated for 10 times with different training/testing splits and the averaged performance is reported.
\textbf{PRID} \cite{hirzer2011person} extracts pedestrian images from recorded trajectory video frames. It has two camera views, each contains 385 and 749 identities, respectively. Only 200 identities appear in both views. In each of 10 single-shot data split, 100 out of that 200 people are chosen randomly for
training, while the remaining 100 of one view are used as the probe
set, and the remaining 649 people's images of the other view are used
as gallery, which thus includes the 100 people in the probe
set.
\textbf{CUHK01} \cite{li2013locally} contains 971 individuals
captured from two camera views. There are two settings; the first is the single-shot setting, that is, one image for each individual in each camera view is randomly selected for both training and testing, and 485 identities are used for training and the other 486 for testing. Under the other setting only 100 identities are used for testing with the rest 871 for training. We use both settings under the supervised setting and only the first setting is used under the unsupervised setting for fair comparisons with the published results.

\noindent \textbf{Evaluation metrics} \quad We use the Cumulated Matching Characteristics
(CMC) curve to evaluate the performance of Re-ID methods. Due to space limitation and for
easier comparison with published
results, we only report the cumulated matching accuracy
at selected ranks in tables rather than plotting the actual
curves. Note that we also use mean average precision (mAP) as suggested
in ~\cite{zheng2015scalable} to evaluate the performance on Market-1501.

\vspace{-0.5cm}
\subsection{Implementation Details}
\vspace{-0.2cm}
We use the Caffe \cite{jia2014caffe} framework to implement our models. In this section we will give some implementation details on input data organisation, detailed structure of the verification and classification subnets and training settings.

\noindent \textbf{Input data organisation\quad} As described above, our network has two different tasks/training objective: the ID classification task and pairwise verification task. There are different ways to organise the training images into minibatches for model training. The simplest way is to organise the training images into pairs. Specifically, one could randomly select positive and negative image pairs and pack them into one minibatch. However, this is very inefficient -- GPU memory is often the hardware bottleneck limiting the number of pairs one could include in each minibatch. To overcome this problem, we follow the minibatch generation scheme introduced in  \cite{ding2015deep} which organises the minibatch according to person identities and generates pairs dynamically. In particular, we keep only one set of base network parameters in the GPU memory and organise our minibatches as follows: In each iteration, we randomly select $K$ person; for each person we then randomly select $M$ images. These $K*M$ distinct images are loaded to form one minibatch. For pair generation,  we first exhaustively generative all the positive and negative pairs according to person identity and then randomly duplicate the positive pairs till the numbers of the positive and negative pairs are equal, i.e., balanced. In this way, much more image pairs can be generated in each minibatch for better training of the model. In our experiments, we randomly select 32 people in each mini-batch, and two images for each person, resulting in 3,968 positive and negative pairs being generated respectively.

\noindent \textbf{Verification subnet\quad} As shown in Fig.\ref{fig:the_network}, each pair of images, after going through the GoogeLeNet base network and pairwise-consistent dropout, are represented by two 1,024D vectors. Inside the verification subnet, they are first subject to  an element-wise subtraction to produce a single 1,024D vector. After passing through a ReLU layer, this vector is then fed into a 1024-dimensional FC layer, followed by a two-node softmax layer.

\noindent \textbf{Classification subnet\quad} The classification subnet consists of a single $N$ nodes softmax layer where $N$ is the unique person identities in the training set.

\noindent \textbf{Auxiliary losses\quad} The original GoogLeNet \cite{szegedy2015going} has another two auxiliary losses/branches extended from the middle layers of the network.  We follow this design pattern by adding extra ID classification and pairwise verification subnets on the two extended branches. This results in a total 6 losses in our network.

\noindent \textbf{Training setting\quad}  The initial learning rate is set to 0.001 and is multiplied by 0.1 every 40K iterations. For supervised two-stepped transfer learning from ImageNet to large Re-ID datasets(CUHK03 and Market-1501), the network is trained for 20K and 150K iterations for each step, respectively. To perform two-stepped transfer learning from large to small Re-ID datasets (e.g.~VIPeR), we train the network for 20K iterations for each step.

\noindent \textbf{Data augmentation\quad} To reduce overfitting, we also perform data augmentation on the Re-ID datasets as in most deep Re-ID works. Similar to \cite{ahmed2015improved}, for each training image, we generate 5 augmented images around the image center by performing random 2D transformation.

\noindent \textbf{Parameter Settings\quad}
%The base network (as well as all insights on base network structure given in our paper) are selected according to experiment results on CUHK03 validation set.
For training our supervised models, the weight between the verification loss and classification loss is 3:1. For our unsupervised co-training method, there is one free parameter $\lambda$ (see Eq.\ref{eq:standardSC}) which needs to be determined. This is done by cross-validation using half of the training data as the validation set.

For all other details about the model architecture and training, please see the source code to be released soon.
%We use Caffe \cite{jia2014caffe} framework to implement our method. For supervised learning on large datasets, we randomly select 32 image pairs in each mini-batch,  The network is trained for 150K iterations. The initial learning rate is set to 0.001 and is multiplied by 0.1 every 40K iterations. To perform supervised two-staged transfer learning, we train the network for 20K iterations for each stage with learning rate set to 0.001. Due to the space limit we can not cover all details, we will release the network model as well as training code that contain all details.
\vspace{-0.5cm}
\subsection{Supervised Transfer Learning}
\label{sec:experiment:supervised}
\vspace{-0.2cm}
\noindent \textbf{Results on large datasets} \quad On the two large Re-ID datasets, namely CUHK03 and Market, one-staged fine-tuning is employed in our model, that is, pretraining on ImageNet (ILSVRC 2012) followed by two-stepped fine-tuning detailed in Sec.~\ref{sec:crossdomain:supervised}. The results of our model are compared with the state-of-the-art deep and non-deep Re-ID models  in Table \ref{tab:sec:experiment:largescale:cuhk03} and Table ~\ref{tab:sec:experiment:main:largescale:market1501} respectively (they are grouped together in the tables). Due to space limit, only the most competitive ones since 2015 are chosen.  We can make the following observations: (1)  Our model significantly outperforms the state-of-the-art: on CUHK03, the gap is 10.1\% using the manually cropped images and 16.0\% using the detected ones. The gap is even bigger for Market, particular on the mAP metric: 26.0\% over Gated S-CNN \cite{varior2016gated} under the single query setting. (2) The best competitors on these two large datasets are all deep learning based. However, their advantages over the hand-crafted feature based models are modest (especially on Market) and far less pronounced than what is widely observed in other visual recognition tasks. This is because the large datasets are still relatively small to release the full potential of a deep model. However, with our model, the gap is clear now. The main reason, as we explained earlier, is that our model is able to transfer feature representations learned from ImageNet thanks to the selected base network (GoogLeNet) and the training objectives (classification + verification loss). In contrast, none of the compared models transfer knowledge from other auxiliary sources -- we found that they cannot even if they are pretrained on ImageNet.

\begin{table}
%\scriptsize
\small
	\begin{center}
		\begin{tabular}{|p{3cm}<{\centering}|p{1.0cm}<{\centering}|p{1.0cm}<{\centering}|}
			\hline
			& \multicolumn{1}{c|}{Manual}  & \multicolumn{1}{c|}{Detected} \\
			\hline
			%Rank & 1 & 5 & 10 & 20 & 1 & 5 & 10 & 20\\
			\hline
			XQDA \cite{liao2015person} & 52.2 & 46.2 \\
			MLAPG \cite{Liao_2015_ICCV} & 57.9 & 51.1 \\
			DNS \cite{zhang2016learning} & 62.5 & 54.7 \\
            LSSCDL \cite{zhangsample} & 57.0 & 51.2   \\
            Siamese LSTM \cite{varior2016siamese} & -   & 57.3  \\
			\hline
			%FPNN \cite{li2014deepreid} & 20.6 & 19.8  \\
			IDLA \cite{ahmed2015improved} & 54.7  & 44.9\\
            DGD \cite{xiao2016learning} & 75.3  & -  \\
            Gated S-CNN \cite{varior2016gated} & -  & 68.1  \\
            EDM \cite{ShiYZLLZL_ECCV16} & 61.3  & 52.0\\
            Joint Learning \cite{wangjoint} & -  & 52.1\\
            CAN \cite{liu2016end} & 65.7 & 63.1 \\
            \hline
			Ours & \bf{85.4} & \bf{84.1} \\
			\hline
		\end{tabular}
	\end{center}
		\caption{Supervised results (Rank 1 matching accuracy in \%) on the CUHK03 dataset.  `-' means no reported result is available.}
			\label{tab:sec:experiment:largescale:cuhk03}
\end{table}

\begin{table}
\small
	\setlength{\belowcaptionskip}{-50pt}
	\setlength{\abovecaptionskip}{-90pt}
	\begin{center}
		\begin{tabular}{|c|cc|cc|}
			\hline
			 & \multicolumn{2}{c|}{Single query}  & \multicolumn{2}{c|}{Multi-query} \\
			\hline
			 & R1 & mAP & R1 &  mAP\\
			\hline
			\hline
			%KISSME (LOMO) \cite{koestinger2012large} & 40.50 & 19.02 & - & - \\
			%MFA (LOMO) \cite{xiong2014person} & 45.67 &  18.24 & - & - \\
%			kLFDA (LOMO) \cite{xiong2014person} & 51.37 &  24.43 & 52.67 & 27.36 \\
            XQDA  \cite{liao2015person} & 43.8 &  22.2 & 54.1 & 28.4 \\
            SCSP \cite{chen2016similarity} & 51.9 & 26.3 & - & - \\
            DNS \cite{zhang2016learning} & 61.0 & 35.6  & 71.5 &  46.0 \\
            Siamese LSTM \cite{varior2016siamese} & - & - & 61.6 & 35.3 \\
			\hline
            Gated S-CNN \cite{varior2016gated} & 65.8 & 39.5 & 76.0 & 48.4 \\
            CAN \cite{liu2016end} & 48.2 & 24.4 & - & - \\
			\hline
			Ours & \bf{83.7} & \bf{65.5} & \bf{89.6} & \bf{73.8} \\
			\hline
		\end{tabular}
	\end{center}
		\caption{Supervised results on Market-1501}
	\label{tab:sec:experiment:main:largescale:market1501}
\end{table}

\begin{table}
	\small
	\begin{center}
		\begin{tabular}{|c|ccc|}
			\hline
			\multirow{2}{*}{} & \multirow{2}{*}{VIPeR} & \multirow{2}{*}{PRID} & CUHK01 \\
            & & & ($N_t$=871/485) \\
			\hline
			\hline
           % XQDA \cite{liao2015person} &  40.0 &  - & - \\
           % MLAPG \cite{Liao_2015_ICCV} &  40.7 & - & - \\
            SCSP \cite{chen2016similarity} & 53.5 & - & - \\
            LSSCDL \cite{zhangsample} & 42.6 & - & -\\
            TMA \cite{martinel2016temporal} & 43.8 &  - & - \\
            $\ell$1 GL \cite{kodirov2016person} & 41.5 & 30.1 & -/50.1 \\
            Siamese LSTM \cite{varior2016siamese} & 42.4 & - & - \\
            \hline
			%Mid-level+LADF \cite{zhao2014learning} & 43.3 & - & -  \\
			Metric Ensemble \cite{paisitkriangkrai2015learning} & 45.9 & - & - \\
            DNS \cite{zhang2016learning} & 51.1 & 40.9 & -/69.0 \\
			\hline
            IDLA \cite{ahmed2015improved} & 34.8 & - & 65.0/47.5\\
            DGD \cite{xiao2016learning} &  38.6 & 64.0* & -/66.6\\
            MCP-CNN \cite{cheng2016person} & 47.8 & 22.0 & -/53.7 \\
            Gated S-CNN \cite{varior2016gated} & 37.8 & - & - \\
            EDM \cite{ShiYZLLZL_ECCV16} & 40.9  & - & 86.6/- \\
            Joint Learning \cite{wangjoint} & 35.8  & - & 72.5/- \\
            CAN \cite{liu2016end} & - & - & 81.0/- \\
			\hline
			Ours & {\bf 56.3} & {\bf 43.6} & {\bf 93.2} / {\bf 77.0} \\
			\hline
		\end{tabular}
	\end{center}
		\caption{Supervised results on VIPeR, PRID and CUHK01. *The DGD results on PRID were obtained by using 10 times more training images from the original PRID video dataset, giving it a huge unfair advantage. }
	\label{tab:sec:experiment:smallscale:supervised}
\end{table}

\noindent \textbf{Results on small datasets\quad} On the three smaller datasets, two-staged transfer learning are required, i.e., ImageNet $\rightarrow$ CUHK03+Market $\rightarrow$ VIPeR/PRID/CUHK01. The comparative results are presented in Table~\ref{tab:sec:experiment:smallscale:supervised}. Note that the compared hand-crafted feature based models have two sub-groups: those with one type of feature and those using multiple based model fusion/ensemble. In addition, most compared deep models use transfer learning, but one-staged (typically from CUHK03+Market) and one-stepped fine-tuning. It can be seen that our deep Re-ID model achieves the best results on all three datasets. The improvements on the two smallest, VIPeR and PRID, are around 3\%, but on the larger CUHK01, the gap is remarkable. In contrast, the existing deep Re-ID models struggle on the small datasets, and none of them can beat the best hand-crafted features based models. This is again due to their inferior transfer learning ability.

\begin{table}
	\small
	\begin{center}
		\begin{tabular}{|c|cc|cc|}
			\hline
			& \multicolumn{2}{c|}{Single query}  & \multicolumn{2}{c|}{Multi-query} \\
			\hline
			 & R1 & mAP & R1 &  mAP\\
			\hline
			\hline
    		SID & 76.6 & 51.7 & 83.6 & 62.2 \\
            PV & 74.6 & 55.0 & 81.5 & 63.1 \\
            TL & 63.3 & 41.5 & 72.4 & 49.7 \\
            SID + PV & {\bf 83.7} & {\bf 65.5} & {\bf 89.6} & {\bf 73.8} \\
            SID + TL & 80.4 & 59.3 & 86.1 & 67.8 \\
            PV + TL & 71.0 & 52.4 & 79.0 & 60.7 \\
            SID + PV + TL & 83.1 & 65.1 & 88.7 & 73.0 \\
			\hline
		\end{tabular}
	\end{center}
	\caption{Comparing different loss selections on Market1501.}
	\label{tab:sec:experiment:largescale:losstype}
\end{table}

\noindent \textbf{Loss selection \quad} We start our ablation study by first examining the selection of losses. We argue that it is the combination of ID classification loss and pairwise verification loss that enables our model to effectively transfer useful representations from the classification-oriented Imagenet and adapt it to the verification task of Re-ID. To validate this claim, we consider three losses: softmax ID classification  (SID), pairwise verification
%\footnote{Note that there are many other forms of PV losses (e.g.~ contrastive loss) apart from the element-subtraction + softmax one used in our model (see Sec.~\ref{sec:model}), but we found that they are less effective.}
(PV)  and triplet loss (TL), and their combinations. All three have been used in existing Re-ID models, but never before has SID been combined with PV. We use the same base network pretrained on ImageNet and test on Market.  We can draw the following conclusions from the results in Table~\ref{tab:sec:experiment:largescale:losstype}: (1) When used alone, SID and PV perform similarly with TL being the worst; (2) When SID is used together with PV or TL, the performance improves dramatically. But without SID, PV+TL gives worse result than PV alone. This suggests clearly that having the classification loss is indeed the key for knowledge transfer from ImageNet. (3) When all three losses are  combined the performance is slightly worse which means that with SID and PV, the TL loss is redundant.

\noindent \textbf{Pairwise-consistent dropout and two-stepped  fine-tuning \quad} Two other contributions made in our model is the pairwise-consistent dropout and the two-stepped fine-tuning. Table \ref{tab:sec:experiment:largescale:dropout} shows that the pairwise-consistent dropout brings about 3\% improvement on both Market-1501 and VIPeR. Note that triplet loss (TL) also benefits, albeit to a smaller extent. We expect whenever these two losses are used, this pairwise-consistent dropout should be chosen over the standard random dropout.
Table \ref{tab:sec:experiment:smallscale:supervised:finetune} suggests that the two-stepped fine-tuning is even more critical, bringing in about 8.7\% at Rank 1 on  VIPeR.

\begin{table}
	\small
	\begin{center}
		\begin{tabular}{|c|c|c|c|c|}
			
			\hline
			Loss  Type & Dropout Strategy & Market-1501 & VIPeR \\
			\hline \hline
            \multirow{2}{*}{SID + PV} & Random & 80.8 & 53.1 \\
            & Pairwise-consistent & {\bf 83.7} & {\bf 56.3}\\
			\hline
            \multirow{2}{*}{SID + TL} & Random  &79.3 & 51.9\\
            & Pairwise-consistent & {\bf 80.4} & {\bf 52.3} \\
			\hline
		\end{tabular}
	\end{center}
	\caption{Rank-1 results of different dropout strategies}
	\label{tab:sec:experiment:largescale:dropout}
\end{table}

\begin{table}
%\scriptsize
\small

	\begin{center} \begin{tabular}{|c|cccc|}
			\hline
			%$Dataset \; Version$ & \multicolumn{4}{c|}{$Manual$}  & \multicolumn{4}{c|}{$Detected$} \\
			 & R1 & R5 & R10 & R20\\
            \hline
            \hline
			One-stepped & 47.6 & 77.2 & 86.8 & 93.1 \\
            \hline
            Two-stepped& {\bf 56.3} &{\bf 83.3} &{\bf 90.5} &{\bf 96.0} \\
			\hline
		\end{tabular}
	\end{center}
		\caption{Two-stepped vs. one-stepped fine-tuning VIPeR}
	\label{tab:sec:experiment:smallscale:supervised:finetune}
\end{table}

\vspace{-0.3cm}
\subsection{Unsupervised Transfer Learning}
\label{sec:experiment:smallscale:supervised}
\vspace{-0.2cm}
Our co-training based unsupervised transfer learning model is compared against the best reported results on the three small datasets
in Table \ref{tab:sec:experiment:smallscale:unsupervised}. Note that to the best of our knowledge, no published deep Re-ID model has attempted this challenging setting. The results clearly show that we can beat the existing hand-crafted features based models by big margins. Compared with the supervised learning results in Table \ref{tab:sec:experiment:smallscale:supervised}, our unsupervised model is very competitive, beating most of them, particularly the deep learning based ones. This indicates that with the developed  unsupervised deep learning model, we can readily deploy a Re-ID system to a new camera network requiring only some person detections but no manual labelling. This is thus a significant step towards real-world deployment of automated Re-ID.

Our ablation study shows that the two models employed in the co-training framework: a soft-label self-training deep model and a discriminative subspace learning model are both effective and co-training yield clear improvements. In addition we compare the proposed model with existing deep unsupervised transfer learning models such as \cite{Ganin_ICML15} to demonstrate that our model is far more effective.

%Specifically, we compare our methods with the following alternatives: the dictionary learning with iterative laplacian regularisation(DLLR) method proposed in \cite{kodirov2015dictionary}, the cross dataset transfer learning(CDTL) method proposed in \cite{pengunsupervised}, the $\ell$1 graph learning($\ell$1 GL) method proposed in \cite{kodirov2016person}. To prove the effectiveness of our proposed soft label method and co-training method, we also give several other results. These are (1) Performance of our base network after training on ImageNet, Market-1501 and CUHK03(denoted as Ours\_BaseNet); (2) Performance of CDTL method using features extracted by Ours\_BaseNet(denoted as CDTL\_OurFeat) and (3) Performance of proposed soft label method(denoted as Ours\_SoftLabel). The result shows that (1) Our base network is very strong as it can produce robust deep features on small datasets, which again prove the conclusion that models pretrained with large auxiliary datasets from other fields can improve the generalisation ability; (2) When trained with the same image features, the proposed soft label methods achieve slightly better performance compared to the state-of-the-art unsupervised Re-ID models(i.e CDTL); (3) The proposed co-training method performs better than all competitors and can beat most supervised models.

\begin{table}
\small
	\begin{center}
		\begin{tabular}{|c|ccc|}
			\hline
			& VIPeR & PRID & CUHK01\\
			\hline \hline
            DLLR \cite{kodirov2015dictionary} & 29.6 & 21.1 & - \\
			CDTL \cite{pengunsupervised} & 31.5 & 24.2 & 27.1 \\
            $\ell$1 GL \cite{kodirov2016person} & 33.5 & 25.0 & 41.0 \\
			\hline
			Ours & {\bf 45.1} & {\bf 36.2} & {\bf 68.8} \\
			\hline
		\end{tabular}
	\end{center}
	\caption{Unsupervised transfer learning results}
	\label{tab:sec:experiment:smallscale:unsupervised}
\end{table}

\vspace{-0.3cm}
\subsection{Evaluations on Base Network Selection}
\vspace{-0.2cm}
\begin{table}
%\scriptsize
\small
	
	\begin{center} \begin{tabular}{|c|c|c|c|c|c|}
			\hline
			Dataset & Network  & Loss  & I.Net Pre.? & R1 & mAP \\
			\hline
			\hline
            \multirow{6}{*}{Market} & \multirow{4}{*}{DGDNet} & \multirow{2}{*}{SID} & YES & 47.5 & 23.1\\
            & & & NO & 47.8 & 23.8 \\
             \cline{3-6}
            & & \multirow{2}{*}{SID + PV} & YES & 71.3 & 48.9\\
            & & & NO & 82.7 & 63.4 \\
            \cline{2-6}
            & \multirow{4}{*}{GoogLeNet} & \multirow{2}{*}{SID} & YES & 76.6 & 51.7\\
            & & & NO & 55.0 & 31.4 \\
            \cline{3-6}
            & &  \multirow{2}{*}{SID + PV} & YES & 83.7 & 65.5\\
            & & & NO & 68.7 & 45.3 \\
			\hline
            \multirow{4}{*}{VIPeR} & \multirow{2}{*}{DGDNet} & \multirow{2}{*}{SID + PV} & YES & 37.4 & N/A\\
            & & & NO & 51.5 & N/A\\
            \cline{2-6}
            & \multirow{2}{*}{GoogLeNet} & \multirow{2}{*}{SID + PV} & YES & 56.3 & N/A\\
            & & & NO & 37.0 & N/A\\
            \hline
		\end{tabular}
	\end{center}
	\caption{Base network comparison}
	\label{tab:sec:experiment:largescale:dgdvsgooglenet}
\end{table}

 In this experiment we compare our base network, GoogLeNet and the one used in the DGD model \cite{xiao2016learning} which we call DGDNet\footnote{Note that we only use their base network, and do not follow their joint-training + domain guided dropout + individual dataset fine-tuning pipeline. Instead, we use the same one/two-staged and two-stepped fine-tuning transfer learning strategy exactly as our models for fair comparison.}. It is chosen because it is representative of the trending smaller/shallower bespoke Re-ID network and has obtained the best results among existing deep Re-ID models. Apart from the base network, the other part of the models are identical, i.e., one-staged transfer learning + two stepped fine-tuning for large Re-ID datasets and  two-staged transfer learning + two stepped fine-tuning for small Re-ID datasets. Table \ref{tab:sec:experiment:largescale:dgdvsgooglenet} shows that: (1) With only SID loss, the smaller base network performs much worse on Market with or without pretraining on ImageNet. However, with the SID+PV combination, the results with DGDNet are much improved, but transfer learning from Imagenet now has a negative effect. (2) With GoogLeNet as base network, transfer learning from ImageNet becomes crucial -- it is too big to be trained from any Re-ID dataset from scratch. (3) On the small VIPeR dataset, with our two-staged transfer learning, the two base networks have quite different behaviours: With the large Market+CUHK03 as auxiliary dataset, our model with DGDNet as base network is quite effective provided no pretraining on ImageNet is conducted, but not as effective as the GoogLeNet base network with ImageNet pretraining (51.5\% vs 56.3\%) -- this shows the advantage of using a deeper base network, that is, it can learn more generalisable feature representations that can benefit small Re-ID datasets. In summary, this results shows that even for smaller deep networks tailor-made for Re-ID, combining classification loss with verification loss is hugely beneficial; but a better network design would be adopting a base network tailor-made for ImageNet and using ImageNet as auxiliary dataset for transfer learning.
 
\vspace{-0.3cm}
\subsection{Alternative Unsupervised Transfer Learning Models}
\vspace{-0.2cm}
We first examine the effectiveness of the co-training strategy. Our co-training model alternates between a soft-label self-training deep model and a graph-regularised subspace learning model. Table \ref{tab:sec:experiment:smallscale:unsupervised:othermethods} shows that both models are effective on its own and when combined in our co-training framework, boost the performance by 2-3\%.

\begin{table}[h!]
	\small
	\begin{center}
		\begin{tabular}{|c|cccc|}
			\hline
			& R1 & R5 & R10 & R20\\
            \hline \hline
            %BaseNet & 37.6 & 63.9 & 73.8 & 83.9 \\
			%\hline
            Self-training & 42.8 & 66.9 & 77.3 & 85.9\\
            Subspace & 42.3 & 71.5 & 79.8 & 87.5 \\
            \hline
			AE  & 36.4 & 62.3 & 74.0 & 81.9\\
			Adversarial \cite{Ganin_ICML15} & 22.8 & 38.6 & 50.3 & 63.9 \\
			\hline
			Ours & {\bf 45.1} & {\bf 73.1} & {\bf 81.7} & {\bf 89.4}\\
			\hline
		\end{tabular}
	\end{center}
	\caption{Evaluations on alternative unsupervised model on VIPeR.}
	\label{tab:sec:experiment:smallscale:unsupervised:othermethods}
\end{table}

In addition, we compare our model with two alternative unsupervised transfer learning methods. The first one combines a CNN with an autoencoder. Autoencoders (AE) \cite{hinton2006reducing} is widely used in unsupervised learning and can be stacked on top of a CNN model to turn it into an unsupervised model. In our CNN+autoencoder model, the input layer of the autoencoder is the feature output of the base network; the middle layer dimension is set to 512 and the output layer has the same dimension as the input layer (1,024). Formally, for each input image $\textbf{x}_i$ in the target dataset $T$, the autoencoder is learned to minimise the following objective:
\begin{equation}
\begin{aligned}
    J(\textbf{x}_i) = \frac{1}{2}||\phi(\textbf{x}_i) - f_d(f_e(\phi(\textbf{x}_i)))||_2^2
\label{eq:2}
\end{aligned}
\end{equation}
Where $f_e$ and $f_d$ denote the mapping functions of the encoder and decoder respectively. Note that since the size of $T$ is too small to train the AE from scratch, we initialize the parameters of the AE layers by first pretraining them using images in the source dataset $S$.
%During testing the encoded representation $f_e(\phi(\textbf{x}_i))$ is computed for each test image $\textbf{x}_i$.
%From Table \ref{tab:sec:experiment:smallscale:unsupervised:othermethods} we can see that the idea of reconstructing images features does not work well since no cross camera constrain is introduced in Eq.\ref{eq:2}.
The second model compared is the deep unsupervised domain alignment model using gradient reversal \cite{Ganin_ICML15}. Specifically, we add a domain classifier connected to the
feature extractor (i.e.~our base network) via a gradient reversal layer that multiplies the gradient by a certain negative constant during the back-propagation based training. The results in Table \ref{tab:sec:experiment:smallscale:unsupervised:othermethods} show that both the compared models yield much weaker performance than the proposed co-training based model. The autoencoder model is weaker because it is not discriminative.  The gradient reversal based models fare even worse, which suggests that the domain adaptation problem for Re-ID poses unique challenges that cannot be addressed by simple domain alignment.

 %In all methods, the based network(denoted as BaseNet) has been pretrained on CUHK03+Market. It can be seen  in Table \ref{tab:sec:experiment:smallscale:unsupervised:othermethods} that none of these alternative deep unsupervised transfer learning methods are as effective as our co-training based one.

\vspace{-0.3cm}
\subsection{Qualitative Results}
\vspace{-0.2cm}
\begin{figure*}[htb]
	\centering
	\includegraphics[width = 1.0\linewidth]{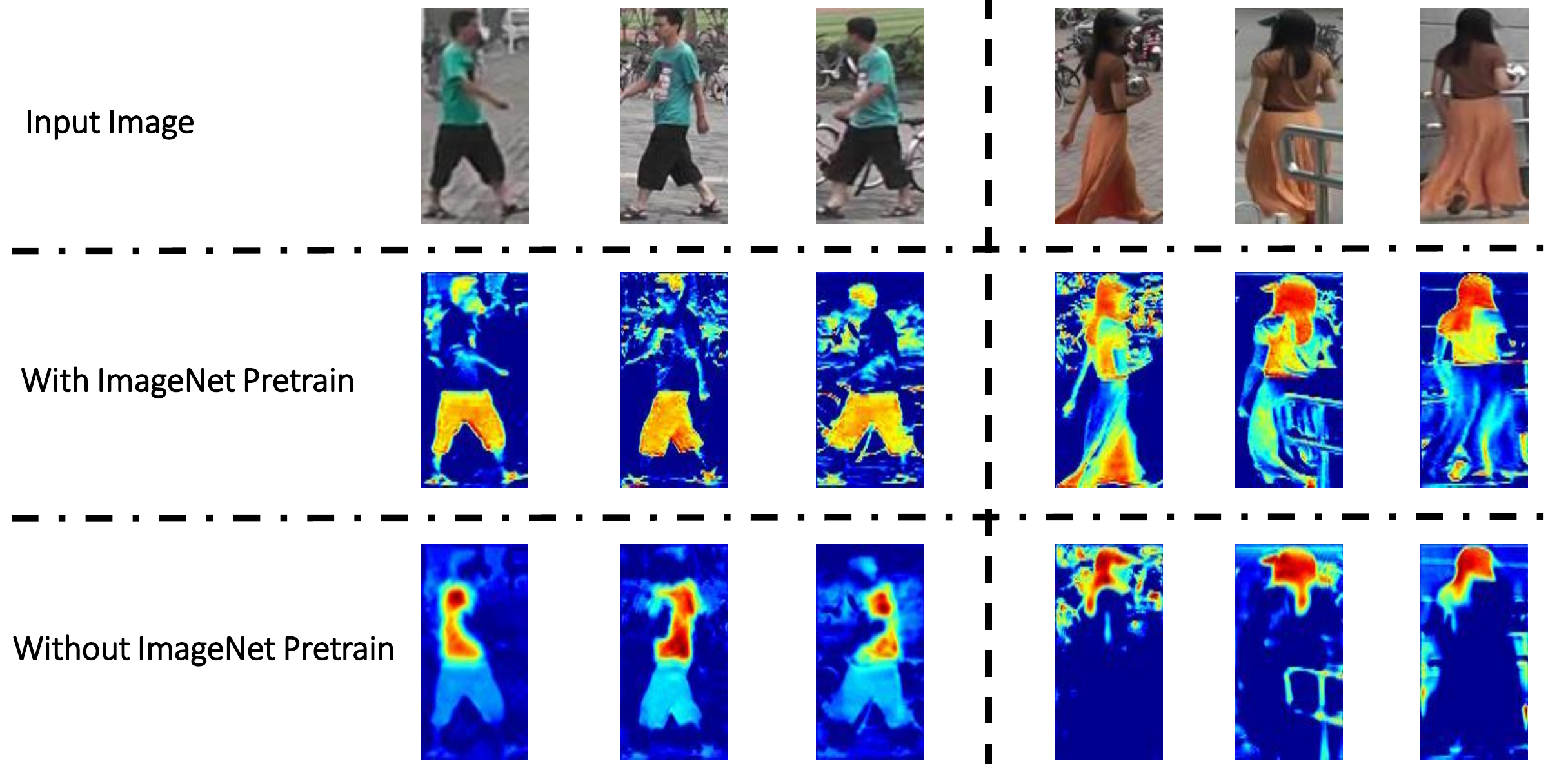}
	\caption{Visualisation of feature responses of different networks. Higher responses are indicated by warmer colours.}
	\label{fig:exp:visualisation}
\end{figure*}
To gain some insights into what the model has actually learned and the contribution of knowledge transfer from large auxiliary dataset such as ImageNet, we visualise in Fig.~\ref{fig:exp:visualisation} some feature responses at the first convolution layer of our GoogLeNet base network which is trained on Market-1501 using the proposed pipeline. For comparison, we also visualise the feature responses of the same layer of another GoogLeNet base network. The only difference between these two networks is that the second one is trained from scratch rather than using ImageNet pretrained parameters. In particular, the first row of Fig.~\ref{fig:exp:visualisation} shows the original input images of two people under different camera views, whilst the second and the third row shows the corresponding feature responses of the two models.

It can be seen clearly that the learned features by the ImageNet pretrained model fire accurately at specific body parts. In contrast, the features learned by the network without ImageNet pretraining are much more fuzzy. This suggests that one of the key benefits of pretraining on Imagenet is that the model is more aware of the concept of visual objects and thus is able to delineate object (person) and object parts (e.g. head, torso, arms etc.) more accurately, which lays a solid foundation for discovering discriminative features for matching people.

%\subsection{Illustrations of Each Components}
%I don't quite understand how we can add illustrations, are we going to show feature maps? Anyway I will follow the instructions later...
\vspace{-0.3cm}
\section{Conclusion}
\label{tab:sec:conclusion}
\vspace{-0.2cm}
We have proposed a number of novel deep transfer learning models to tackle the challenging person Re-ID problem with small datasets. Our experiments validated the claim that using a deep base network together with a combination of classification and verification loss is key for transferring representations learned from large image classification datasets. Importantly, we show for the firs time that, a co-training based deep unsupervised transfer learning model can perform effective Re-ID without any labelled data.

{\small
\bibliographystyle{ieee}
\bibliography{egbib}
}

\end{document}